\documentclass[11pt]{article}

\usepackage[]{EMNLP2022}

\usepackage{times}
\usepackage{latexsym}
\usepackage{graphicx} 
\usepackage{amsmath}
\usepackage{bm}
\usepackage{subfigure}
\usepackage{amsmath}
\usepackage{amsfonts}

\usepackage[noend]{algpseudocode}
\usepackage{algorithmicx,algorithm}
\usepackage[T1]{fontenc}

\usepackage[utf8]{inputenc}

\usepackage{microtype}

\usepackage{inconsolata}

\title{Curriculum Learning Meets Weakly  Supervised Modality Correlation Learning}

\author{Sijie Mai \and Ya Sun \and Haifeng Hu \\
        School  of Electronics and Information Technology, Sun Yat-sen University \\ \texttt{\{maisj,suny278\}@mail2.sysu.edu.cn}, \ \ \texttt{huhaif@mail.sysu.edu.cn}}

\begin{document}
\maketitle
\begin{abstract}
In the field of multimodal sentiment analysis (MSA), a few studies have leveraged the inherent modality correlation information stored in samples for self-supervised learning. However, they feed the training pairs in a random order without consideration of difficulty. Without human annotation, the generated training pairs of self-supervised learning often contain noise. If noisy or hard pairs are used for training at the easy stage, the model might be stuck in bad local optimum. In this paper, we inject curriculum learning into weakly supervised modality correlation learning. The weakly supervised correlation learning leverages the label information to generate scores for negative pairs to learn a more discriminative embedding space, where negative pairs are defined as two unimodal embeddings from different samples. To assist the correlation learning, we feed the training pairs to the model according to difficulty by the proposed curriculum learning, which consists of elaborately designed scoring and feeding functions. The scoring function computes the difficulty of pairs using pre-trained and current correlation predictors, where the pairs with large losses are defined as hard pairs. Notably, the hardest pairs are discarded in our algorithm, which are assumed as noisy pairs. Moreover, the feeding function takes the difference of correlation losses as feedback to determine the feeding actions (`stay', `step back', or `step forward'). The proposed method reaches state-of-the-art performance on MSA.
\end{abstract}

\section{Introduction}
With the rapid development of social media, multimodal data have been widely-used to perform many downstream tasks, including multimodal sentiment analysis \cite{MAG-BERT}, multi-omics integrative analysis\cite{mvib}, human action recognition\cite{human_action}, etc. Multimodal sentiment analysis, aiming to mine humans' sentiments and opinions from language, audio, and visual sequences, has attracted significant attention.

\begin{figure}[h]
  \centering
  \includegraphics[width=1.0\linewidth]{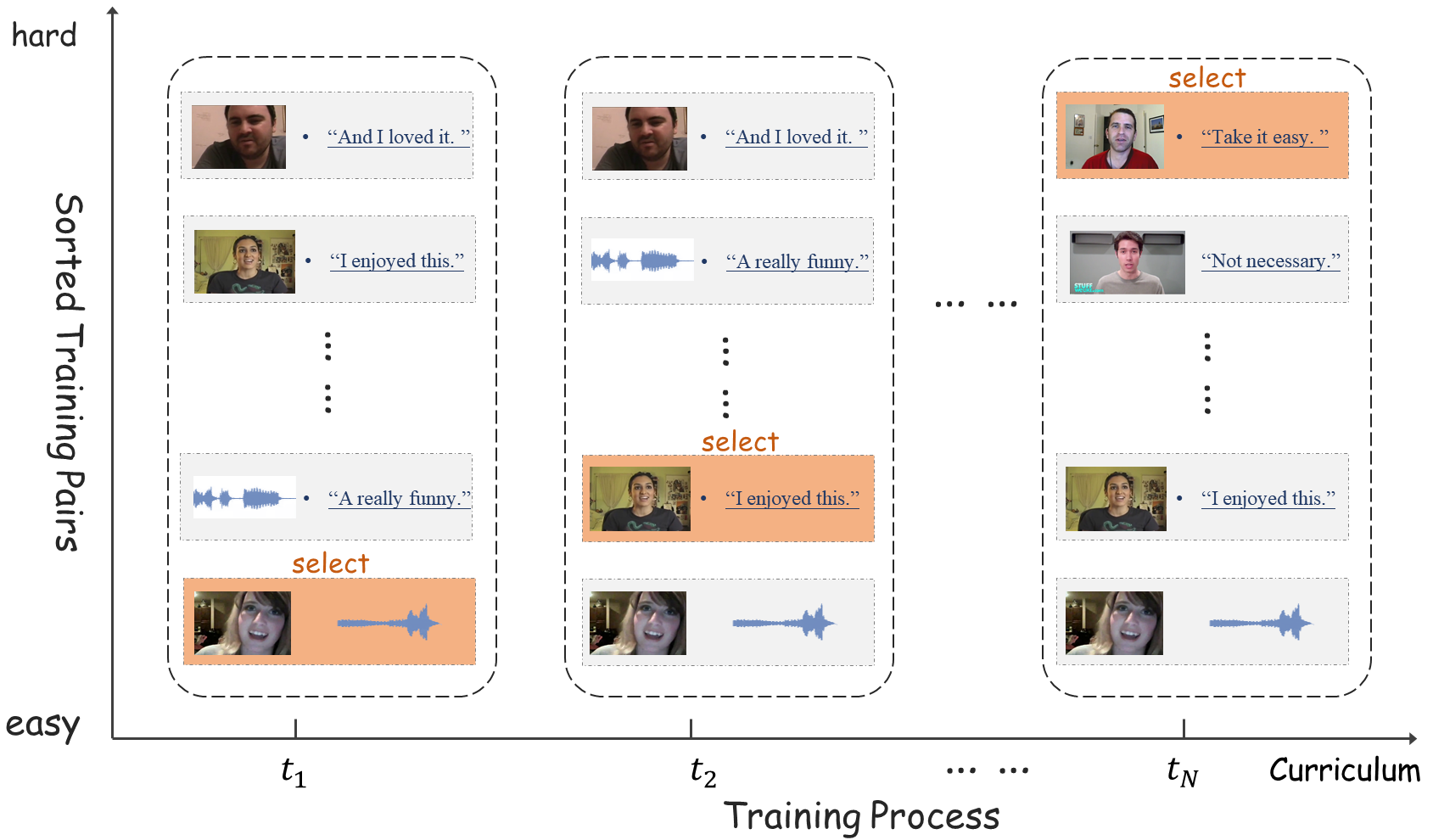}
  \caption{\label{sample}\textbf{Illustration of the curriculum learning. } }
\vspace{-0.1cm}
\end{figure}


Manual annotation of large-scale multimodal datasets is challenging and time-consuming \cite{amrani2021noise}. Many models suffer from overfitting due to limited training samples and the large amount of parameters introduced by complex fusion algorithms\cite{Zadeh2017Tensor,MULT} or large pre-trained transformers\cite{BERT,XLNet}, which degrades the generalizability. 
Recently, to leverage the correlation information between modalities inherently stored in each sample, a few contrastive learning frameworks are proposed \cite{hycon,COBRA,contrastive_multiview,robust_contrastive_multi_view}, which generate abundant training pairs of unimodal embeddings across samples to train a robust model. These methods learn the similarity between modalities, such that unimodal embeddings from the same sample are pulled closer while those from different samples or classes are pushed away. 
However, they feed the generated pairs to the model in a random order without consideration of their difficulty, which might cause inaccurate learning. If hard pairs are used for training at early stage where the model is not well trained, the optimization may be stuck in a bad local optimum \cite{multimodal_curriculum}. More severely, without human supervision, the training data of unsupervised or weakly supervised tasks often contain noise. For instance, two modalities from the same sample might convey contradictory semantic meanings. On the other hand, two unimodal pairs can share semantic similarity even if they come from different samples.
The noisy pairs are often the hardest pairs that cannot be well predicted by the model, which should be identified to prevent the feature extraction capability from being damaged.

In contrast, the education of humans is highly organized, which is based on a curriculum that introduces different concepts at different times according to difficulty, exploiting previously learned concepts to ease the learning of new abstractions. Inspired by the mode of human education,
we focus on selecting the generated pairs for training dynamically (see Fig.~\ref{sample}), and follow the idea of curriculum learning which enables training data to be fed into the model in a certain order\cite{curriculum_power,bengio2009curriculum,curriculum_survey2}. By choosing which data to present and in which order to present them, we can guide the learning and find a better local minimum for the model.

Specifically, we first construct a weakly supervised modality correlation learning task to capture the shared information between modalities, where a positive pair and a negative pair consists of unimodal representations from the same sample and from different samples respectively. The task is said to be weakly supervised as it incorporates label information to score the pairs to learn a more discriminative embedding space for downstream tasks (while the score is not always correct).
Then, we inject curriculum learning into correlation learning which consists of difficulty scoring and pair feeding functions.
The scoring function determines the hardness of training pairs based on pre-trained and current correlation predictors, and sorts the training pairs based on hardness. The feeding function determines the pace by which the training pairs are fed to the model. Different from traditional methods\cite{curriculum_power,multimodal_curriculum}, our feeding function has three candidate actions (`stay', `step back', or `step forward'), which enables a more refined and accurate feeding strategy. It takes the difference of current and previous correlation losses as feedback to determine the feeding actions. The proposed curriculum learning enables the model to be learned gradually from easy pairs to hard pairs which effectively prevents the model from falling into bad local optimum. To address the issue of noisy pairs, we dynamically identify and discard the hardest pairs during training to minimize the negative effect of noisy data.


In brief, the contributions are listed as follows:

\begin{itemize}
\item We construct a Weakly Supervised Correlation Learning framework and innovatively equip the framework with Curriculum Learning (WSCL-CL) that enables the model to progressively learn from easy pairs to hard pairs. In this way, we can prevent the model from falling into bad local optimum.
\item In curriculum learning, difficulty scoring function is introduced to sort the training pairs based on pre-trained and current predictors, and feeding function is elaborately designed to determine the pace by which the training pairs are fed to the model. Moreover, we address the noisy pair problem during training which is not considered in previous methods\cite{COBRA,hycon}.
\item The proposed method achieves state-of-the-art performance on multimodal sentiment analysis. We show that the introduced curriculum learning can improve the performance of the weakly supervised learning.
\end{itemize}

\section{Related Work}

With the availability of multimodal data, MSA draws increasing attention for its capability to interpret human language and mine sentiments and opinions. Many works focus on designing fusion strategies to obtain informative multimodal embedding \cite{Poria2017A}. Two simple and explicit fusion methods are early fusion (i.e., feature-level fusion) \cite{Wollmer2013YouTube,Poria2017Convolutional,Poria2017Context} 
and late fusion (i.e., decision-level fusion)\cite{Nojavanasghari2016Deep,Personality,tac_late_fusion}.
Recently, modern fusion methods are proposed to explore intra-/inter- modal interactions. Specifically, tensor-based methods \cite{T2FN,Zadeh2017Tensor} draw increasing attention because they can learn joint embedding with high expressive power. 
Graph-based fusion methods\cite{ARGF,MOSEI,graph_fusion2} learn interactions on the graph across time series and modalities. To highlight important information across modalities, many algorithms apply cross-modal attention mechanisms \cite{Zadeh2018Multi, MULT,Multilogue-Net}. To improve the interpretablility of multimodal fusion, quantum-inspired fusion \cite{Quantum} and capsule-based fusion \cite{MRM,our_capsule} are proposed. 
Moreover, a few methods use KL-divergence, canonical correlation analysis and so on to regularize the learning of unimodal distributions and better match different modalities \cite{ICCN,nips_cca,nll,sun_meta}.
With the success of BERT\cite{BERT}, there has been a trend to fine-tune pre-trained transformers using multimodal data\cite{CM-BERT,MAG-BERT,MISA}. However, these methods do not fully leverage the modality correlation information inherently stored in each sample, and may easily suffer from overfitting and have poor generalizability especially when the  dataset is small\cite{mcl}.

More recently, self-supervised learning on multimodal data has attracted significant research attention, where the self-supervised learning tasks typically include modality matching (especially image-text pair) and masked reconstruction \cite{vision_language_adversarial,vision_language_vilbert,vision_language_lxmert,mcl}. Another popular strategy of self-supervised learning is to apply contrastive learning to learn representations by pushing unimodal representations
describing the same sample closer, and pushing those of different samples apart \cite{trimodal_contrastive_tricolo,COBRA,contrastive_comir,robust_contrastive_multi_view}.
For example,
HyCon \cite{hycon} explores intra-modal and inter-modal dynamics across samples, which defines two unimodal representations from the same class as positive pairs. However, these methods do not consider feeding the training pairs into the model based on difficulty, which might cause the model to be stuck in the bad local optimum. In contrast, we perform weakly supervised modality correlation learning to learn a more discriminative embedding space for downstream task, and leverage curriculum learning to enable a more smooth and accurate learning. Moreover, we also seek to identify noisy modality pairs for minimizing the negative effect of noisy data. Curriculum learning has been applied in various tasks and shows promising results \cite{curriculum_power,curriculum_survey,multimodal_curriculum,curriculum_facial,hybrid_curriculum,bengio2009curriculum}.  To the best of our knowledge, we are the first to introduce curriculum learning into weakly supervised/unsupervised multimodal learning algorithms.

\section{Algorithm}

\subsection{Notations and Task Definitions}
\begin{figure*}
\setlength{\belowcaptionskip}{-0.3cm}
\centering
\includegraphics[scale=0.31]{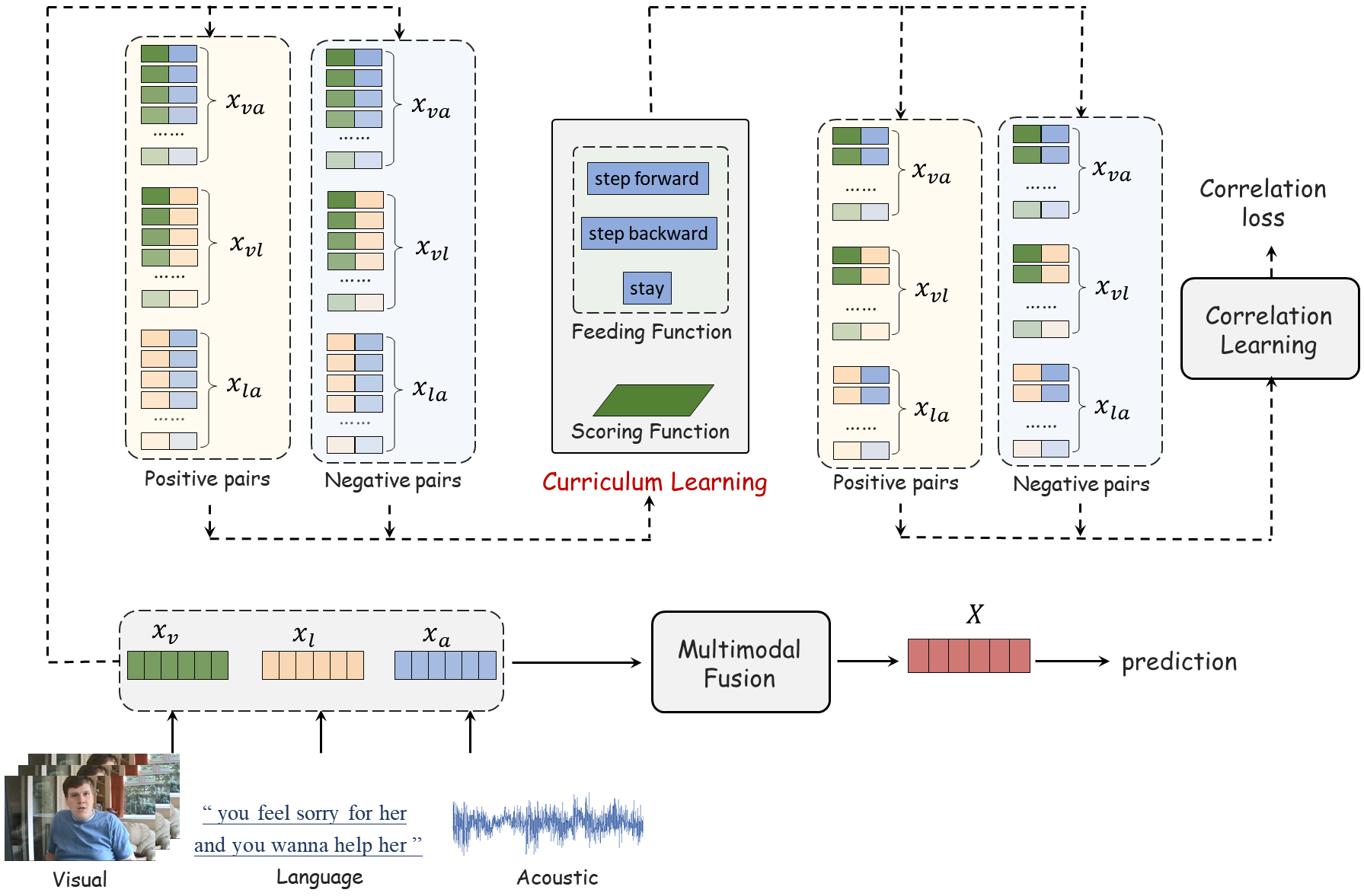}
\caption{\label{1}\textbf{The diagram of the proposed WSCL-CL.} 
}
\end{figure*}

Our downstream task is multimodal sentiment analysis (MSA). The input of MSA is an utterance \cite{Olson1977From}, i.e., a segment of a video bounded by a sentence. Each utterance has three modalities, i.e., acoustic ($a$), visual ($v$), and language ($l$). The input sequences of acoustic, visual, and language modalities are denoted as $\bm{U_a} \in \mathbb{R}^{T_a \times d_a}$, $\bm{U_v} \in \mathbb{R}^{T_v \times d_v}$, and $\bm{U_l} \in \mathbb{R}^{T_l \times d_l}$ respectively, where $T_m$ and $d_m$ are the sequence length and feature dimensionality respectively ($m\in \{a, v, l \}$). MSA aims to predict the sentiment score  based on three unimodal sequences. After being processed by unimodal learning network $F^m$, the input sequence $\bm{U_m} \in \mathbb{R}^{T_m \times d_m}$ becomes the unimodal representation $\bm{x_m} \in \mathbb{R}^{d}$. For conciseness, the structures of $F^m$ are illustrated in the Appendix. The pipeline of our algorithm is illustrated in Fig.~\ref{1}. We first introduce the modality correlation learning task, and then elaborate on our curriculum learning.

\subsection{Modality Correlation Learning Task}

Formally, let $\bm{x}_{m_1}^j$, $\bm{x}_{m_2}^j$, ..., $\bm{x}_{m_k}^j$ denote the representations of $k$ modalities describing the same sample $j$ ($k=3$ in the case of MSA), where $j\in \{1,..,n\}$ and $n$ is the batch size. 
Correlation learning aims to enable unimodal networks to extract high-level semantic representations and capture the shared information across modalities within each sample\cite{mcl}. We take modality $m_i$ and modality $m_j$ as an example to illustrate the pipeline. Firstly, to construct the negative pairs, we sample $N$  unimodal representations from modality $m_i$ and modality $m_j$ respectively, where $N$ is the number of negative pairs. Then we concatenate the representations of each two modalities to generate negative bimodal representations:
\begin{equation}
\setlength{\abovedisplayskip}{1pt}
\setlength{\belowdisplayskip}{1pt}
\begin{aligned}
   \bm{x}^{o}_{m_im_j} =
   \bm{x}_{m_i}^{o_1}\oplus \bm{x}_{m_j}^{o_2},\ \ o_1\! \neq\! o_2
\end{aligned}
\end{equation}
where $\oplus$ denotes concatenation operation, $ \bm{x}^{o}_{m_im_j}\in \mathbb{R}^{2 \cdot d}$ ($o \in \{1,2,..,N\}$) is a negative bimodal representation whose constructed modalities come from different samples. $N$ is set to $\beta \cdot n$, where $\beta$ is the negative sampling factor.
By sampling abundant modality pairs for training, the negative effect of small dataset can be mitigated.  
Similarly, we construct positive bimodal representations as:
\begin{equation}
\setlength{\abovedisplayskip}{1pt}
\setlength{\belowdisplayskip}{1pt}
\begin{aligned}
  \! \bm{x}^{o}_{m_im_j}\!\!=\! \bm{x}_{m_i}^{o_1}\!\oplus\! \bm{x}_{m_j}^{o_2},\  o_1\! =\! o_2,\ o\! \in\! \{1,2,..,n\}
\end{aligned}
\end{equation}

To determine the score of the generated pairs, we first calculate the distance between the labels of the associated samples:
\begin{equation}
\setlength{\abovedisplayskip}{1pt}
\setlength{\belowdisplayskip}{1pt}
\begin{aligned}
   d^o = \sqrt{(\bar{l}_{o_1}- \bar{l}_{o_2})^2}
\end{aligned}
\end{equation}
The final correlation score is defined as:
\begin{equation}
\label{score}
\begin{aligned}
   y^o =
   \begin{cases}
   \frac{1}{d^o+\gamma}, & o_1 \neq o_2 \\
   1, & o_1 = o_2
   \end{cases}
\end{aligned}
\end{equation}
where $\bar{l}_{o_1}$ and $\bar{l}_{o_2}$ are the true labels of the sample $o_1$ and $o_2$ respectively, which are continuous values that indicate the intensity of sentiment. 
$\gamma$ is a hyperparameter that determines the maximum score of negative pairs ($\gamma>1$). We assign a score within zero and one to each modality pair. If two unimodal representations are from the same object, $y^o$ is equal to 1. If they come from different objects but share the same label, we have $y^o=\frac{1}{\gamma}$. And if they have different labels, we have $y^o<\frac{1}{\gamma}$. Consequently,  the more semantic similarity the two modalities share, the larger $y^o$. Label information of each object is considered here to learn a more discriminative feature space for downstream task (that is why we call it weakly supervised).

A correlation predictor is then leveraged to predict the correlation score of each training pair, and then the correlation loss is calculated based on the predicted score:
\begin{equation}
\begin{aligned}
\label{eq_c}
   s^o = CP(\bm{x}^{o}_{m_im_j};\ \bm{W}_{cp}) =  \bm{W}_{cp} \bm{x}^{o}_{m_im_j} 
\end{aligned}
\end{equation}
\begin{equation}
\label{eq_l}
\begin{aligned}
  \ell_{m_i m_j}\!\! =\!\frac{1}{2N_p}\!\sum_{o=1}^{N_p} |s^o\! -\! y^o|\! +\! \frac{1}{2N_n}\sum_{o=1}^{N_n}\! |\hat{s}^o\!\!-\! \hat{y}^o|
\end{aligned}
\end{equation}
\begin{equation}
\label{eq_l2}
\begin{aligned}
  \ell_c = \frac{1}{C_k^2} \sum_i^{i< j} \sum_{j=2}^k \ell_{m_i m_j},\ \ \ i, j \in \{1,2,..,k\}
\end{aligned}
\end{equation}
where $C_k^2$ is the number of bimodal correlation losses, $\ell_c$ is the overall correlation loss, $CP$ denotes the correlation predictor,  $\bm{W}_{cp}\in \mathbb{R}^{1\times 2d}$ is the parameter matrix of the correlation predictor, $s^o$ denotes the predictive score of the $o^{th}$ positive pair, and $\hat{s}^o$ is the score of negative pair (here we add the hat to identify positive and negative pairs). Note that $N_p$ and $N_n$ are the number of selected positive and negative pairs respectively (determined by the curriculum learning). In Eq.~\ref{eq_l}, mean absolute error (MAE) is used to calculate the learning loss. By minimizing the correlation loss of positive pairs, the model can learn to identify whether two modalities are describing the same object, encouraging the model to capture shared information  across various modalities within each object.  
Via minimizing the correlation loss of negative pairs, the model can learn to discover the distinguishable information of each sample with respect to other samples. 
Instead of directly using dot product or distance metric to calculate the similarity of unimodal embeddings as in contrastive learning \cite{hycon,COBRA,trimodal_contrastive_tricolo,triplet}, correlation learning utilizes a learnable predictor which can explore complex correspondence relationship between modalities\cite{mcl}. 


\subsection{Curriculum Learning}
The curriculum learning determines how to choose the training pairs presented to the model and in which order to present them\cite{bengio2009curriculum}. In this way, we can guide the learning of  modality correlation learning task and find a better local minimum for the model. The core and the main innovation of our curriculum learning lies in the design of difficulty scoring function and pair feeding schedule,  which are introduced below. Notably, the design of the curriculum learning is independent of concrete learning tasks. Other than the correlation learning task, the proposed curriculum learning is applicable to any tasks as long as the self-/weakly-supervised learning loss is given.

\subsubsection{Difficulty Scoring Function}
Scoring Function is designed to determine the difficulty of the training pairs. 
In this paper, we propose to score the pairs by leveraging the pre-trained and current predictors. Compared to inner product or Euclidean distance that cannot reveal high-level correlation, our method is more target-oriented.

To identify the hardness of the pairs, we first pre-train a correlation predictor $CP_{pre}$ on the same dataset. At the warm-up training stage, we only use the pre-trained $CP_{pre}$ to determine the difficulty. For a batch of training pairs $\{ \bm{x}_{m_im_j}^o | o = 1,2,... \}$, we feed them into $CP_{pre}$, which outputs the correlation scores  $\{ s_{m_im_j}^o | o = 1,2,... \}$. We then compute the correlation loss of each training pair $\{ l_{m_im_j}^{o'} | o = 1,2,... \}$,  and finally sort the pairs by their loss values (which are defined as the difficulties of the pairs). However, as the training strategies of $CP_{pre}$ and current $CP$ are different, the $CP_{pre}$ might not serve as an excellent difficulty scoring function. Therefore, after the warm-up training stage where the current $CP$ becomes discriminative, we incorporate the predictive loss of $CP$ to generate difficulty score:
\begin{equation}
\label{eq_l2}
\begin{aligned}
  s_d^o = l_{m_im_j}^{o'} +\lambda \cdot l_{m_im_j}^{o},\ \ \ o \in \{1,2,..\}
\end{aligned}
\end{equation}
where $l_{m_im_j}^{o}$ is the correlation loss produced by current $CP$, $s_d^o$ denotes the difficulty score after warm-up training stage, and $\lambda$ is set to 0.8 in our experiment.
The sorted training pairs are then sent into the feeding function.

\subsubsection{Pair Feeding Function}
The feeding function is designed to determine by which the training data are fed to the model. In this paper, we elaborately design the feeding function according to the feedback of the correlation loss to adaptively change the feeding strategy.

Different from previous methods\cite{curriculum_power,bengio2009curriculum,multimodal_curriculum}, our feeding function has three actions to be taken at each iteration, namely, `step forward', `step backward', and `stay'. Firstly, we divide the sorted training pairs ($\{ \bm{x}_{m_im_j}^o \}$ for short) into $c$ partitions based on difficulty. The feeding function is responsible for choosing one partition for training at each iteration, where the pairs from the same partition are assumed to have approximately the same difficulty. Notably, for positive and negative pairs, the value of $c$ can be different to ensure a more flexible choosing mechanism.  Generally, we follow the idea that if the model learns fast from the tasks of a particular difficulty (i.e., the correlation loss decreases intensely), we should sample the tasks of the same difficulty at the next iteration (i.e., `stay') to enable the model to be learned fast. In contrast, if the model learns little from the tasks of a particular difficulty for a certain number of iterations, we should `step forward' and train the model with more challenging tasks. Finally, if the performance drops considerably, we should `step backward' to let the model learn more from easy pairs and become more expressive before feeding harder pairs. The choosing procedure is shown in Algorithm 1. The feeding function is run for $2\cdot C_k^2$ times to select positive/negative unimodal pairs.

As shown in Algorithm 1, we introduce the counting factor $count$ and patience factor $p$ to enable a more stable training. Only when $count$ is larger or equal to $p$ which implies that the model learns little from the tasks of current difficulty and becomes steady during training, can we feed more difficult tasks to the model. In this way, we can prevent the correlation loss from oscillating. Notably, as presented in Line 5 of Algorithm 1, when the model is already trained by the hardest partition, we randomly sample other pairs for training, which prevents the catastrophic forgetting of previous knowledge.

Moreover, as elaborated in Introduction section, without human supervision, the generated training data of unsupervised or weakly supervised tasks often contain noise. To minimize the negative effect of noisy data, we assume the hardest pairs (with correlation loss greater than 95\% of all pairs) as noisy and discard them during training.

\begin{algorithm}[t]
\caption{Pair Feeding Function}
\raggedright
\label{alg1}
\textbf{input:} Patience factor $p$, counting factor $count$, forward factor $f$, backward factor $b$, number of partitions $c$, training pairs $\{\bm{x}_{m_im_j}^o \}$, previous choosing index $c_i$, previous correlation loss $\ell_{m_i m_j}^{pre}$ \\
\textbf{output:} A partition of training pairs $\{\bm{x}_{m_im_j}^{o'} \}$\\
    \begin{algorithmic}
        \State 1. Compute current correlation loss $\ell_{m_i m_j}^{now}$ using all training pairs in $\{\bm{x}_{m_im_j}^o \}$. 
		\State 2. Divide $\{\bm{x}_{m_im_j}^o \}$ into $c$ partitions.
		\State 3. If ($\ell_{m_i m_j}^{now}$ - $\ell_{m_i m_j}^{pre}$)/ $\ell_{m_i m_j}^{pre}$ > $b$:\\
  \	\ \ \ \ \ \ \ $count = 0$ \\
	\ \ \ \ \ \ \ \ if $c_i>1$:\\
	\ \ \ \ \ \ \ \ \ \ \ \ \ \  $c_i = c_i - 1$ \ \ \ \%step backward\\
\ \ \ \   elseif ($\ell_{m_i m_j}^{pre}$ - $\ell_{m_i m_j}^{now}$)/ $\ell_{m_i m_j}^{pre}$ > $f$:\\
 \ \ \ \ \ \ \ \ $count = 0$\ \ \ \ \ \ \%stay\\
\ \ \ \  elseif $count >= p$:\\
		  \ \ \ \ \ \ \ \ $count = 0$ \\
		\  \ \ \ \ \ \ \ if $c_i<c$:\\
		  \ \ \ \ \ \ \ \ \ \ \ \ \ \  $c_i = c_i + 1$\ \ \ \%step forward\\
	\ \ \ \	   else:\\
		\  \ \ \ \ \ \ \ $count = count + 1$\ \ \ \ \%stay.
		 \State 4. Choose the  $(c_i)^{th}$ partition $\{\bm{x}_{m_im_j}^{o'} \}$.
		 \State 5. if $c_i==c$:\\
		 \ \ \ \ \ \ \ Randomly sample other training pairs  and\ \ \ \ \ \ \ put them into $\{\bm{x}_{m_im_j}^{o'} \}$.
		 \State 6. $\ell_{m_i m_j}^{pre}\longleftarrow \ell_{m_i m_j}^{now}$.
    \end{algorithmic}
\end{algorithm}

\subsection{Sentiment Prediction}
Our algorithm is independent of the concrete fusion method.
In practice, we apply a deep neural network on the concatenated unimodal representations to learn  high-level interactions between modalities:
\begin{equation}
 \bm{X} = DNN(\bm{x}_a\oplus\bm{x}_v\oplus\bm{x}_l;\ \theta_{DNN})
\end{equation}
where $DNN$ consists of several fully connected layers. We then learn a multimodal predictor that is composed of fully connected layers to generate the sentiment prediction based on $\bm{X}$:
\begin{equation}
   \label{eq66}
  \tilde{l} = MP(\bm{X};\ \theta_{MP})
\end{equation}
where $\tilde{l}$ is the predicted label, and $MP$ is the multimodal predictor.
We use mean square error (MSE) to compute the task loss:
   \begin{equation}
   \label{eq77}
  \mathcal{L} = (\bar{l}-\tilde{l})^2
\end{equation}
where $\bar{l}$ is the ground truth label,  and $\mathcal{L}$ is the MSE for prediction. 
The total loss is weighted sum of the task loss and correlation loss:
  \begin{equation}
   \label{eq99}
  \mathcal{L} \longleftarrow \mathcal{L} + \alpha \cdot \ell_c
\end{equation}
where $\alpha$ is the weight of the correlation loss $\ell_c$. 

\section{Experiments}
In this section, we conduct extensive experiments to evaluate the effectiveness of the proposed WSCL-CL. \textbf{For the lack of space, more details about the introduction of baselines and the feature extraction details are placed in the Appendix. }
\subsection{Datasets}

1) \textbf{CMU-MOSI}\cite{Zadeh2016Multimodal} is a widely-used dataset for multimodal sentiment analysis from online websites, which contains a collection of more than 2k video clips. Each video clip is annotated with sentiment on a [-3,3] Likert scale, where +3 denotes the strongest positive sentiment and -3 the strongest negative. To be consistent with prior works \cite{sun_meta,MAG-BERT}, we use 1,281 utterances for training, 229 utterances for validation, and 685 utterances for testing.

2) \textbf{CMU-MOSEI}\cite{MOSEI} is a large dataset of multimodal sentiment analysis. The dataset consists of more than 20k utterances from more than 1,000 YouTube speakers, covering 250 distinct topics. All the utterances are randomly chosen from various topics and monologue videos. 
In our algorithm, we use the sentiment label of CMU-MOSEI to perform sentiment label, where the range of the sentiment label is the same as that of the CMU-MOSI. Following previous works \cite{sun_meta,MAG-BERT,hycon}, we use 16,265 utterances as training set, 1,869 utterances as validation set, and 4,643 utterances as testing set.


\subsubsection{Evaluation Metrics}

We adopt the following metrics to evaluate the performance of the model: 1) Acc7: 7-way accuracy, sentiment score classification; 2) Acc2: binary accuracy, positive or negative; 3) F1 score; 4) MAE: mean absolute error and 5) Corr: the correlation between the model's prediction and human evaluation. Notably, to compute the 7-way accuracy, we round up the model prediction to an integer from -3 to 3. 
Note that following \cite{MAG-BERT}, when calculating Acc2, F1 score, corr, and MAE, we do not use the neutral utterances.

\subsection{Experimental Settings}\label{sec:exper_detail}


We develop our model with the PyTorch framework on GTX2080Ti on CUDA 10.1 and torch version 1.4.0. The proposed model is trained using Adam \cite{Kingma2014Adam} optimizer. The detailed hyperparameter setting can be referred to Table~\ref{t2223}. Notably, the negative sampling factor $\beta$ is set to be larger than that in the pre-training stage, but we only sample a part of the generated pairs for training via the pair feeding function.

\begin{table}[t]
\centering
 \caption{ \label{t2223}Hyperparameters of WSCL-CL.}
 \resizebox{.95\columnwidth}{!}{\begin{tabular}{c|c|c}
 \hline
     & CMU-MOSI & CMU-MOSEI  \\
 \hline
  Batch Size  &  64 & 48  \\
   Learning Rate  & 2e-5  & 1e-5  \\
   Weight $\alpha$  & 1  & 1  \\
   Training Epochs &  20 & 50  \\
   Dimensionality $d$ &  100 & 100  \\
  Negative Sampling Factor $\beta$ (pre-training)  & 4  &  4  \\
  Negative Sampling Factor $\beta$ & 30  &  30  \\
  Scale Factor $\gamma$  & 1.4  &  1.4 \\
  patience Factor $p$ & 6 & 5  \\
  forward Factor $f$ & 0.1 & 0.1  \\
  backward Factor $b$ & 0.15 & 0.15  \\
   $c$ for negative pairs & 10 & 10  \\
   $c$ for positive pairs & 2 & 8  \\
 \hline
 \end{tabular}}
\end{table}%

\begin{table}[t]
\centering
 \caption{ \label{t1}\textbf{ The comparison with baselines on CMU-MOSI.}  The best results are highlighted in bold, and the second best results are marked with underlines. The results of ICCN \cite{ICCN} are copied from the original paper since the codes are unavailable. }
\resizebox{.99\columnwidth}{!}{\begin{tabular}{c|c|c|c|c|c}
 \hline
   & Acc7 ($\uparrow$) & Acc2 ($\uparrow$) & F1 ($\uparrow$) & MAE  ($\downarrow$) & Corr ($\uparrow$) \\
 \hline
TFN-BERT \cite{Zadeh2017Tensor} & 44.7 & 82.6  & 82.6 & 0.761 & 0.789 \\
 LMF-BERT \cite{Liu2018Efficient} & 45.1 &  84.0 & 84.0 & 0.742 & 0.785  \\
  MFN-BERT \cite{Zadeh2018Memory} & 44.1 &  83.5 & 83.5 & 0.759 & 0.786 \\
 MULT-BERT \cite{MULT} & 41.5 &  83.7 & 83.7 & 0.767 & \textbf{0.799} \\
   GFN-BERT \cite{ARGF} & 47.0 & 84.3 & 84.3 & 0.736 & 0.790 \\
 ICCN-BERT \cite{ICCN} & 39.0 &  83.0 & 83.0 & 0.860 & 0.710 \\
 MAG-BERT \cite{MAG-BERT} & 42.9 & 83.5 & 83.5 & 0.790 & 0.769 \\
TFR-Net \cite{tfr-net} & 42.6 & 84.0 & 83.9 & 0.787 & 0.788 \\
AMML \cite{sun_meta} & 46.3 & 84.9 & 84.8 & 0.723 & 0.792 \\
HyCon\cite{hycon}  & \underline{46.6} & \underline{85.2}  & \underline{85.1} & \underline{0.713} & 0.790 \\
\hline
WSCL-CL & \textbf{47.5} & \textbf{86.3} & \textbf{86.2} & \textbf{0.712} & \underline{0.798} \\
 \hline
 \end{tabular}}
\end{table}%

\begin{table}[t]
\centering
 \caption{ \label{t2}\textbf{ The comparison with baselines on CMU-MOSEI dataset.} }
\resizebox{.99\columnwidth}{!}{\begin{tabular}{c|c|c|c|c|c}
 \hline
    & Acc7 ($\uparrow$) & Acc2 ($\uparrow$) & F1 ($\uparrow$) & MAE  ($\downarrow$) & Corr ($\uparrow$) \\
 \hline

 TFN-BERT \cite{Zadeh2017Tensor} & 51.8 & 84.5   & 84.5 &  0.622 & \underline{0.781} \\
  LMF-BERT \cite{Liu2018Efficient} & 51.2 & 84.2  & 84.3 & 0.612  & 0.779 \\
 MFN-BERT \cite{Zadeh2018Memory} & 52.6 & 84.8  & 84.8 & 0.607 & 0.771 \\
 MULT-BERT \cite{MULT} & 50.7 &  84.7 & 84.6 & 0.625 & 0.775 \\
    GFN-BERT \cite{ARGF} & 51.8 &  85.0& 85.0 & 0.611 & 0.774 \\
  ICCN-BERT \cite{ICCN} & 51.6  &  84.2 & 84.2 & \textbf{0.565} & 0.713 \\
 MAG-BERT \cite{MAG-BERT} & 51.9 & 85.0 & 85.0 & 0.602 & 0.778 \\
 TFR-Net \cite{tfr-net} & 51.7 & 85.2 & 85.1 & 0.606 & \underline{0.781} \\
 AMML \cite{sun_meta} & 52.4 & 85.3 & 85.2 & 0.614 & 0.776 \\
 HyCon\cite{hycon}  & \underline{52.8} & \underline{85.4}  & \underline{85.6} & 0.601 & 0.776 \\
\hline
WSCL-CL & \textbf{53.3} & \textbf{86.1} & \textbf{86.0} & \underline{0.577} & \textbf{0.794} \\
 \hline
 \end{tabular}}
\end{table}%

\subsection{Comparison with Baselines}

In this section, we compare our proposed method with baselines on the task of multimodal sentiment analysis. 
As shown in Table~\ref{t1} and ~\ref{t2}, the contrastive learning based algorithm HyCon performs better than other baselines and sets up a high baseline. Nevertheless, WSCL-CL still outperforms HyCon and obtains the best performance on the majority of the evaluation metrics. Specifically, on CMU-MOSI dataset, WSCL-CL outperforms HyCon by 0.9\% on Acc7, 1.1\% on Acc2 and F1 score. The improvement is remarkable considering that the humans do not always perform well on multimodal sentiment analysis\cite{Zadeh2017Tensor}. On CMU-MOSEI dataset, WSCL-CL yields 0.7\% improvement on Acc2 and 1.2\% improvement on F1 score compared to the current state-of-the-art  HyCon\cite{hycon}. Our method also reaches state-of-the-art performance on Acc7 and Corr. Compared to HyCon\cite{hycon}, WSCL-CL allows the modality-specific information to be preserved in an elegant way, addresses the noisy pair problem, and considers to feed the training pairs according to difficulty to find a better local optimum for the model. 
These results demonstrate the effectiveness of our proposed model, indicating the importance of learning the correlation between modalities and injecting curriculum learning into weakly-supervised learning task.

\subsection{Ablation Study}
In this section, we conduct extensive ablation studies to evaluate the effectiveness of the components in our WSCL-CL:

\textbf{1) Correlation Learning}:
In the case of `W/O Correlation Learning' (see Table~\ref{t3}), we remove the correlation learning task and the accompanied curriculum learning.
As shown in  Table~\ref{t3}, the performance drops dramatically by about 3.5 points in Acc7 and 2.5 points in Acc2 and F1 score, demonstrating the importance of correlation learning to better correlate unimodal representations in a more discriminative embedding space;

\textbf{2) Curriculum Learning}:
In the case of `W/O Curriculum Learning', we remove the curriculum learning but retain the correlation learning. The results suggest that the curriculum learning brings over 1\% improvement  to the multimodal systems on Acc7, Acc2 and F1 score. The performance on other evaluation metrics is also improved. These results reveal that selecting pairs for training dynamically based on difficulty is of great significance to correlation learning, which can find a better local minimum for the parameters of the model;

\textbf{3) Patience Factor}:
In the case of `W/O Patience Factor', the feeding function selects harder pairs immediately if no significant gain is obtained on the easier pairs. The results on all the evaluation metrics decrease considerably. We argue that this is because without the patience factor, the learning of modality correlation suffers from oscillating and a worse local optimum is reached;

\textbf{4) Backward Action}:
In the case of `W/O Backward', we remove the backward action in the feeding function such that the feeding function cannot select easier pairs after harder pairs are used for training.
We observe that Acc7 and Acc2 drop by about 2\% and 1\% respectively, demonstrating the importance of the design of backward action in learning a more expressive model by `looking back' at easier training data;

\textbf{5) Discarding Hardest Pairs}:
The performance on all the evaluation metrics declines when the hardest pairs are used for training. The results suggest that the hardest pairs are harmful to the performance of the model which are often the noisy pairs and needed to be addressed during training;

\textbf{6) Random Sampling}:
In the case of `W/O Random Sampling', we only use the hardest training pairs if the algorithm has already selected the hardest partition of $\{\bm{x}_{m_im_j}^o \}$ for training. The performance of the model slightly drops, which might be due to the catastrophic forgetting of previous knowledge. Although random sampling does not improve the performance by a large margin, it remains a good practice considering that it is simple and introduces no additional parameters. 

\begin{table}[t]
\centering
 \caption{ \label{t3}\textbf{ Ablation studies on the CMU-MOSI dataset.} 
 }
\resizebox{.95\columnwidth}{!}{\begin{tabular}{c|c|c|c|c|c}
 \hline
    & Acc7 & Acc2 & F1 & MAE & Corr \\
 \hline
 W/O Correlation Learning & 43.8 & 83.8 & 83.8 & 0.756 & 0.784 \\
 W/O Curriculum Learning & 46.1 & 84.9 & 84.8 & 0.715  & 0.797 \\
  W/O Patience Factor & 47.0 & 85.7 & 85.6 & 0.728 & 0.793 \\
 W/O `Backward' & 45.4 & 85.5 & 85.3 & 0.714 & 0.797 \\
 W/O Discarding Hardest Pairs   & 46.7 & 85.8 & 85.7  & 0.718 & 0.793 \\
 W/O Random Sampling & 47.3 & 86.1 & 86.1 & \textbf{0.710} & \textbf{0.798} \\
 \hline
WSCL-CL  & \textbf{47.5} & \textbf{86.3} & \textbf{86.2} & 0.712 & \textbf{0.798} \\
  \hline
 \end{tabular}}
\end{table}%


\subsection{Discussion on Difficulty Scoring Function}
We analyze the performance of different kinds of difficulty scoring functions in this section. The compared methods include the widely-used inner product, Euclidean distance, current correlation predictor alone, and pre-trained correlation predictor alone. As shown in Table~\ref{t4}, combining the performance on all the evaluation metrics, the combination of the pre-trained $CP_{pre}$ and current $CP$ gives the best result, which demonstrates the effectiveness of the proposed method in identifying the difficulty of training pairs. 

In addition, all the scoring functions reach remarkable results, indicating that the curriculum learning is widely applicable given reasonable difficulty scoring function and pair feeding function. Euclidean distance achieves the best performance on MAE and Corr, but our method still slightly outperforms it considering that our improvement on Acc2 and F1 score is remarkable. We argue that this is because Euclidean distance scores the pairs by the similarity between two unimodal embeddings which cannot reveal the high-level correlation between different modalities. In contrast, our method utilizes the correlation loss as difficulty score, which is more target-oriented.

Notably, using pre-train $CP_{pre}$ as difficulty scoring function is  time-consuming considering that we have to pre-train the model. Actually, although the current CP might not be discriminative at the start of training, it ends up performing well. The other difficulty scoring functions also reach competitive performance. For the sake of time complexity, one can freely choose the time-efficient difficulty scoring functions listed in Table~\ref{t4}.

\begin{table}[t]
\centering
 \caption{ \label{t4}\textbf{ Discussion on Difficulty Scoring Function.} In the case of `Euclidean distance', we define hard positive (negative) pairs as the pairs with long (short) distance, and `Inner Product' vice verse.}
\resizebox{.95\columnwidth}{!}{\begin{tabular}{c|c|c|c|c|c}
 \hline
    & Acc7 & Acc2 & F1 & MAE & Corr \\
 \hline
 Euclidean distance & 47.5 & 85.5 & 85.4 & \textbf{0.703} & \textbf{0.804} \\
  Inner Product & 46.9 & 85.7 & 85.6 & 0.710 & 0.795 \\
 Pre-trained $CP_{pre}$ & 47.2 & 85.7 & 85.6 & 0.716 & 0.797 \\
 Current $CP$ & \textbf{47.6} & 85.5 & 85.4 & 0.719 &0.792\\
 \hline
$CP_{pre}$ + $CP$  & 47.5 & \textbf{86.3} & \textbf{86.2} & 0.712 & 0.798 \\
  \hline
 \end{tabular}}
\end{table}%

\subsection{Visualization of Feeding Function}
\begin{figure}[h]
  \centering
  \includegraphics[width=1.0\linewidth]{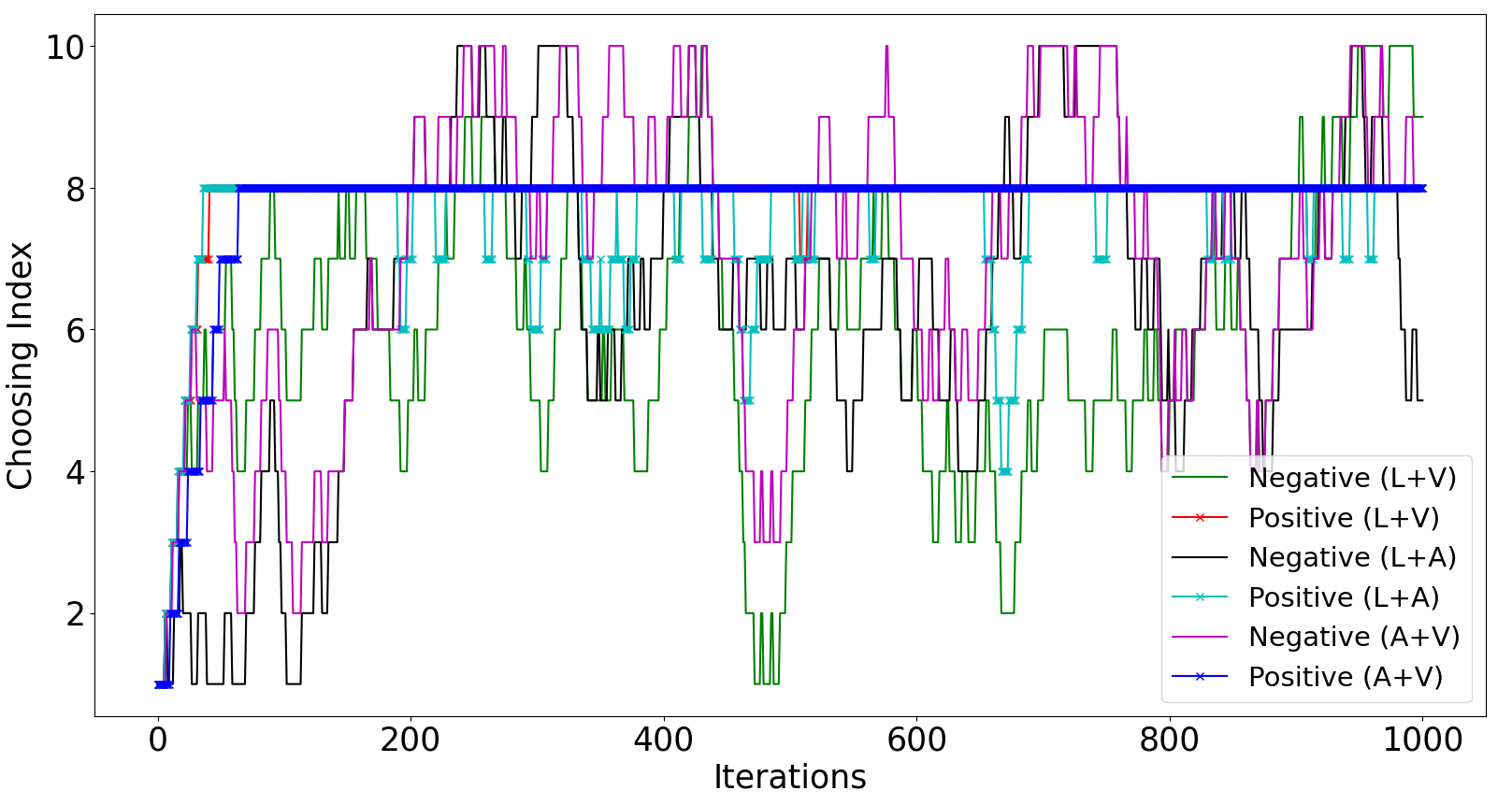}
  \caption{\label{step}\textbf{Visualization of the value of the choosing index $c_i$ for different training pairs. } }
\end{figure}

In this section, we visualize the values of $c_i$ for different positive and negative training pairs on CMU-MOSEI dataset. For CMU-MOSEI, the number of partitions of training pairs is set to 10 and 8 for negative and positive pairs, respectively. As shown in Fig.~\ref{step}, we observe that the learning of positive pairs is much faster than negative pairs. Especially for the positive language-visual (L+V) and acoustic-visual (A+V)  pairs, their choosing index $c_i$ reaches the largest value at the early stage of the training, and almost remains unchanged as training deepens, which indicates that the correlation between these modalities is easy to learn. For positive language-acoustic (L+A) pairs, the choosing index $c_i$ also reaches the largest value at the early stage, but it suffers from oscillating as training deepens, suggesting that the correlation between language and acoustic modalities is hard to learn. We argue that this is because the literal meaning of spoken language and the tone does not always coordinated (e.g., the `tone' modality is often negative but the spoken language is positive literally if sarcastic tone was used). For negative pairs, the trend of the changes with respect to $c_i$ is opposite to that of the positive pairs, that is, they generally suffer from large oscillation. These results suggest that identifying the modalities belonging to the same sample is easy, while identifying modalities from different samples is much more difficult. It indicates that discovering the distinguish information of each sample and  learning the correspondence relationship across different samples are difficult, and points out a possible way to improve our correlation learning: paying more attention to correctly locate different samples in the embedding space.

\subsection{Analysis on Model Complexity}
In this section, we evaluate the time complexity and space complexity of WSCL-CL. For the training time, all the compared baselines are tested under the same GPU device and experimental setting (batch size is consistent across all methods).
Quantitatively, as shown in Table~\ref{t666}, combining the time of pre-training and training, the time consumption of WSCL-CL with pre-trained $CP_{pre}$ is about 524 seconds on CMU-MOSI dataset. Without pre-trained $CP_{pre}$, the time consumption of WSCL-CL (which also achieves competitive performance) is 385 seconds. In comparison, the running time of HyCon, MULT-BERT and MFN-BERT is 440, 378, and 455 seconds, respectively. For space complexity, considering that we use a simple fusion method and the additional trainable parameters brought by WSCL-CL are only the parameters in the correlation predictor, the space complexity is acceptable compared to other baselines using complex fusion structures. Specifically, the number of parameters of WSCL-CL is 109,917,955. For comparison, the number of parameters of HyCon, MULT-BERT and MFN-BERT is 109,209,149, 110,838,399, and 111,256,325, respectively. Overall, the model complexity of WSCL-CL is acceptable.

\begin{table}[!htb]
\centering
\caption{ \label{t666}\textbf{The comparison of model complexity on CMU-MOSI dataset.}}
\resizebox{.95\columnwidth}{!}{\begin{tabular}{c|c|c}
 \hline
    & The number of parameters & Running Time\\
 \hline
MAG-BERT \cite{MAG-BERT} & 110,853,121 & 293\\
TFN-BERT \cite{Zadeh2017Tensor} & 161,409,399&275  \\
GFN-BERT \cite{ARGF} &109,232,350 & 274\\
MFN-BERT \cite{Zadeh2018Memory} &111,256,325 & 455\\
MULT-BERT \cite{MULT} &110,838,399 & 378\\
HyCon \cite{hycon} &109,209,149 & 440\\
\hline
WSCL-CL (W/O $CP_{pre}$) & 109,917,955  & 385 \\
WSCL-CL & 109,917,955 & 524 \\
 \hline
 \end{tabular}}
\end{table}%

\section{Conclusion}
We introduce the curriculum learning and inject it into a weakly supervised modality correlation learning task. Via curriculum learning, we can feed the model with training pairs based on their difficulties, such that the model does not fall into a bad local optimum. The main novelty of this paper is the design of the curriculum learning, in which
the difficulty scoring and pair feeding functions are proposed.
The scoring function computes the difficulty of pairs using  pre-trained and current correlation predictors, 
and the feeding function takes the difference in correlation losses as feedback to determine the feeding action.
The proposed method achieves state-of-the-art performance on multimodal sentiment analysis.

\section{Acknowledgement}

This work was supported by the National Natural Science Foundation of China (62076262, 61673402, 61273270, 60802069)

\section*{Limitations}
The main limitation of this paper is that using the learned correlation predictor might still not be a perfect scoring function. This is because the learning loss of the learned predictor does not approximate zero, and thus the learned predictor might wrongly classify a training pair into `easy' or `hard' pair. Moreover, the training of the pre-trained correlation predictor increases the time complexity.

\bibliography{anthology, sentiment2}
\bibliographystyle{acl_natbib}

\appendix

\section{Appendix}
\label{sec:appendix}

\subsection{Unimodal Networks $F_{m}$}

Due to the excellent capability of  Transformer\cite{transformer} to model long  sequences, we follow HyCon\cite{hycon} to use Transformer-based\cite{transformer} model to extract high-level representations from unimodal sequences. We apply a regular Transformer encoder to extract the acoustic and visual features. 
Following the state-of-the-art algorithms \cite{MISA,MAG-BERT,hycon}, BERT \cite{BERT} is used to extract the high-level language representation. 

Specifically, the procedures of the BERT network are shown as below: 
\begin{equation}
\label{eq6}
\begin{split}
   &\bm{\hat{X}}_{l}=\text{BERT}(\bm{U}_l)\\
    \bm{X}_{l}=&\operatorname{Conv} 1 \mathrm{D}\left( \bm{\hat{X}}_{l}, K_l\right) \in \mathbb{R}^{T_l \times d}\\
\end{split}
\end{equation}
where $\operatorname{Conv} 1 \mathrm{D}$ denotes the temporal convolution with kernel size $K_l$ set to 3, which is used for mapping the output dimensionality of BERT to the shared dimensionality $d$. The final language representation $\bm{x}_{l}$ is defined as representation of the last time step of $\bm{X}_{l}$, for the reason that it incorporates the information from prior time steps during processing. 
For the Transformer encoder network, the procedures are presented as follows:
\begin{equation}
\label{eq6}
\begin{split}
&\bm{\hat{X}}_{m}=\operatorname{Conv} 1 \mathrm{D}\left( \bm{U}_{m}, K_m\right) \in \mathbb{R}^{T_m \times d}\\
   &\bm{X}_{m}=\text{Transformer}(\bm{\hat{X}}_{m}) \in \mathbb{R}^{T_m \times d}
\end{split}
\end{equation}
Here we also take the representation in the last time step of $\bm{X}_{m}$ as the final representation $\bm{x}_{m}$. 
The generated unimodal representation $\bm{x}_{m}$ is used for fusion and multimodal association learning.

\subsection{EXPERIMENT}

\subsubsection{Feature extraction Details}\label{sec:exper_detail}

Facet$^1$ is used for extracting a set of visual features that are composed of facial action units, facial landmarks, head pose, etc. 
COVAREP \cite{Degottex2014COVAREP} is used for extracting acoustic features, which consist of 12 Mel-frequency cepstral coefficients, pitch tracking, speech polarity, glottal closure instants, spectral envelope, etc. 
Following the state-of-the-art methods \cite{MAG-BERT,MISA,CM-BERT,hycon}, BERT \cite{BERT} is used to extract the high-level textual representation. All the baselines use the same feature set in our experiment. For CMU-MOSEI dataset, the input feature dimensionality of language, acoustic, and visual modality are 768, 74, and 35, respectively. For CMU-MOSI, the input feature dimensionality of language, acoustic, and visual modality are 768, 74, and 47, respectively. 

\let\thefootnote\relax\footnotetext{\textsuperscript{\rm 1}iMotions 2017. https://imotions.com/}

\textbf{2) Training details}: We develop our model with the PyTorch framework on GTX2080Ti with CUDA 10.1 and torch version of 1.4.0. Our proposed model is trained using Adam \cite{Kingma2014Adam} optimizer. 

\subsubsection{Baselines}

1) \textbf{Memory Fusion Network} (\textbf{MFN}) \cite{Zadeh2018Memory}, which designs delta-attention module and multi-view gated memory network to explore cross-modal interactions;

2) \textbf{Multimodal Transformer} (\textbf{MULT}) \cite{MULT}, which learns multimodal representation by translating the source modality into the target modality using cross-modal Transformer \cite{transformer}; 

3) \textbf{Graph Fusion Network} (\textbf{GFN})\cite{ARGF}, which designs a graph neural network to explore unimodal, bimodal, and trimodal interactions;

4) \textbf{Tensor Fusion Network} (\textbf{TFN}) \cite{Zadeh2017Tensor}, which applies outer product over unimodal representations to jointly learn unimodal, bimodal, and trimodal dynamics;

5) \textbf{Low-rank Modality Fusion} (\textbf{LMF}) \cite{Liu2018Efficient}, which leverages low-rank weight tensors to deal with the high complexity problem of tensor fusion;

6)  \textbf{Interaction Canonical Correlation Network} (\textbf{ICCN}) \cite{ICCN}, which fuses language features with visual and acoustic features respectively to obtain two bimodal representations, and then applies a canonical correlation analysis (CCA) network on the bimodal representations to generate multimodal representation;

7) \textbf{Multimodal Adaption Gate BERT} (\textbf{MAG-BERT}) \cite{MAG-BERT}, which proposes an attachment  called multimodal adaptation gate derived from RAVEN\cite{RAVEN} that enables large pre-trained transformers to accept multimodal data during fine-tuning; 

8) \textbf{Transformer-based Feature Reconstruction Network} (\textbf{TFR-Net}) \cite{tfr-net}, which proposes a feature reconstruction network to improve the robustness of multimodal model for the random missing in modality sequences;

9) \textbf{Hybrid Contrastive Learning} (\textbf{HyCon}) \cite{hycon}, which explores inter-class relationships and intra-/inter-modal interactions  using intra-/inter-modal contrastive learning and semi-contrastive learning;

10) \textbf{Adaptive Multimodal Meta-Learning} (\textbf{AMML}) \cite{sun_meta}, which applies meta-learning to learn better unimodal representation for multimodal fusion.

\end{document}